\documentclass[journal]{IEEEtran}
\usepackage{cite}

\usepackage[font=small]{caption}
\usepackage{bm}
\usepackage{amssymb}
\usepackage{enumerate}
\usepackage{amsmath}
\usepackage{booktabs}
\usepackage{multirow}
\usepackage{subcaption}
\usepackage{lipsum}
\usepackage{hyperref}
\usepackage{pifont}
\usepackage{dblfloatfix}    
\usepackage[table,xcdraw]{xcolor}

\ifCLASSINFOpdf
   \usepackage[pdftex]{graphicx}
\else
\fi
\hypersetup{
 linkcolor=red,
 citecolor=blue,
 colorlinks=true
}

\begin{document}
%
\title{Looking at the Driver/Rider in Autonomous Vehicles to Predict Take-Over Readiness}
%
%
%

\author{Nachiket~Deo,
       and~Mohan~M.~Trivedi,~\IEEEmembership{Fellow,~IEEE}
\thanks{The authors are with the Laboratory for Intelligent and Safe Automobiles
(LISA), University of California at San Diego, La Jolla, CA 92093 USA}
\thanks{(email:ndeo@ucsd.edu, mtrivedi@ucsd.edu)}
\thanks{}}

%
%

\markboth{}%
{Shell \MakeLowercase{\textit{et al.}}: Bare Demo of IEEEtran.cls for IEEE Journals}
%



\maketitle

\global\csname @topnum\endcsname 0
\global\csname @botnum\endcsname 0

\begin{figure*}
\centering
\includegraphics[width=\textwidth]{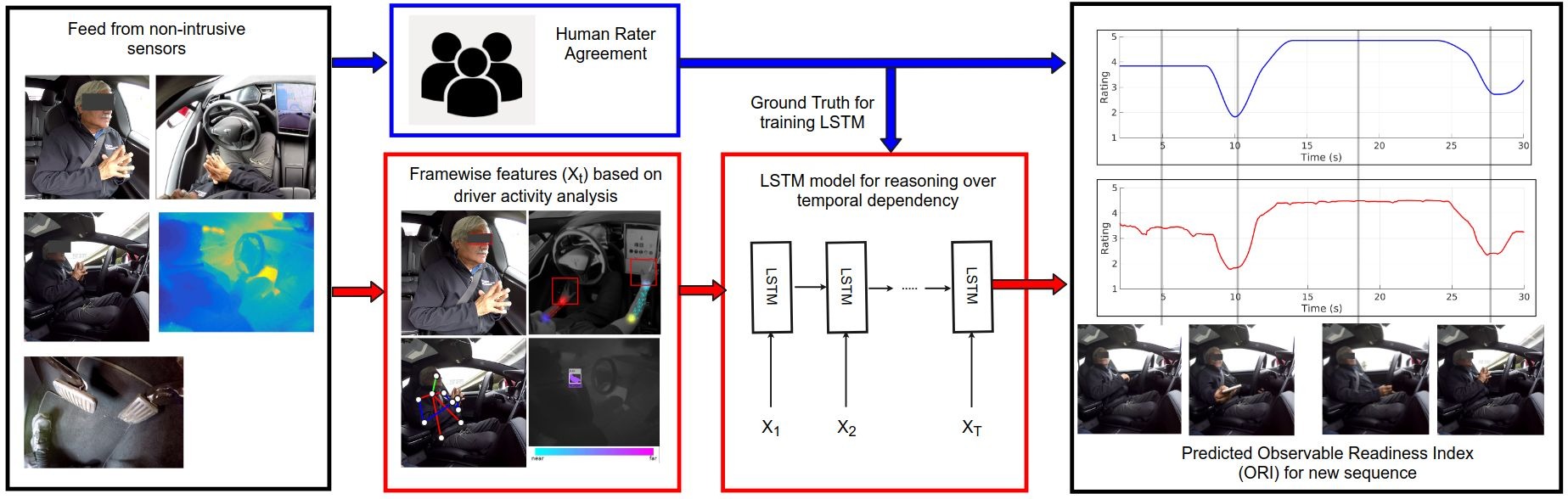}
\caption{\textbf{Overview of our approach:} We wish to continuously estimate the driver's readiness to take-over control from an autonomous vehicle based on feed from vision and depth sensors capturing the driver's complete state. We define a continuous ground truth value for take-over readiness of the driver based on ratings provided by multiple human raters observing sensor feed. We term this the 'Observable Readiness Index (ORI)'.We process the sensor feed frame-by-frame using models for driver activity analysis and propose an LSTM model to learn the temporal dependencies in the frame-wise features. Our model continuously estimates the ORI of the driver.}
\label{fig_pull}
\end{figure*}

\begin{abstract}
Continuous estimation the driver's take-over readiness is critical for safe and timely transfer of control during the failure modes of autonomous vehicles. In this paper, we propose a data-driven approach for estimating the driver's take-over readiness based purely on observable cues from in-vehicle vision sensors. We present an extensive naturalistic drive dataset of drivers in a conditionally autonomous vehicle running on Californian freeways. We collect subjective ratings for the driver's take-over readiness from multiple human observers viewing the sensor feed. Analysis of the ratings in terms of intra-class correlation coefficients (ICCs) shows a high degree of consistency in the ratings across raters. We define a metric for the driver's take-over readiness termed the 'Observable Readiness Index (ORI)' based on the ratings. Finally, we propose an LSTM model for continuous estimation of the driver's ORI based on a holistic representation of the driver's state, capturing gaze, hand, pose and foot activity. Our model estimates the ORI with a mean absolute error of 0.449 on a 5 point scale.

\end{abstract}


%
\IEEEpeerreviewmaketitle

\section{Introduction}

The overarching goal of autonomous vehicle research is the development of fully automated systems capable of driving in any traffic scenario. The occupants of such a vehicle would then be mere passengers, without access to controls. However, to safely develop the technology to achieve this goal, there needs to be shared control between the vehicle and a human driver. This can be seen in the 5 levels of vehicle automation defined by the Society of Automotive Engineers (SAE) \cite{sae2016taxonomy}, with levels 2 to 4 corresponding to some form of shared control. Conditionally autonomous vehicles (level 3), can operate autonomously under specific traffic scenarios like lane keeping on freeways while maintaining a desired speed. Such vehicles are now commercially available \cite{lvl3_1,lvl3_2}. However, a human occupant, behind the wheel, is expected to monitor the automated system and be prepared for \textit{take-over requests}. These are cases where control needs to be transferred from the vehicle to the human during failure modes of the system. Continuous estimation of this occupant's take-over readiness is thus critical for safe and timely transfer of control. In the remainder of this paper, we use the term 'driver' in the context of conditionally autonomous vehicles to refer to the occupant responsible for taking over control from the vehicle.

Prior research \cite{liu2016driver,liang2007real,liang2014hybrid,bergasa2006real,li2015predicting,wollmer2011online} has addressed the closely related problem of estimating \textit{driver distraction} under manual driving conditions. Driver distraction has been defined as the diversion of the driver's attention away from activities critical for safe driving toward a competing activity, which may result in insufficient or no attention to activities critical for safe driving \cite{regan2011driver}. Conditionally autonomous driving raises the possibility of drivers engaging in secondary activities unobserved during manual driving, as well as more freely engaging in previously known secondary activities. While sophisticated computer vision algorithms have been proposed for driver activity analysis \cite{tawari2014robust,tawari2014driver,lee2011real,vasli2016driver,fridman2016driver,fridman2016owl,vora2017generalizing,vora2018driver,ohn2013vehicle,ohn2014beyond,ohn2014hand,borghi2018hands,rangesh2016hidden,deo2016vehicle,molchanov2015hand,yuen2018looking,yuen2016looking,rangesh2018handynet,tran2011pedal,tran2012modeling}, relatively few works \cite{gold2013take,braunagel2017ready,zeeb2015determines} have addressed the problem of mapping driver activity to take-over readiness. This could be attributed to two main challenges. First, there is a lack of naturalistic drive datasets observing driver activity in conditionally autonomous vehicles. A comprehensive naturalistic drive dataset capturing a large range of driver behaviors would allow for data-driven approaches to map driver activity to take-over readiness. Second,
defining the ground truth for take-over readiness is a challenging task. Data-driven approaches hinge on the availability of ground-truth of the quantity being estimated. While electroencephalogram (EEG) sensors allow for the most faithful representation of the driver's brain activity \cite{jung1997estimating,zhang2016vehicle,lin2014wireless}, they are too intrusive to be viable in commercial vehicles. Another approach used in recent studies \cite{braunagel2017ready} is to define take-over readiness based on take-over time and take-over quality in experimental trials with take-over requests issued to drivers performing secondary activities. However, the nature of the task restricts such experiments to simulator settings.

In this paper, we propose a data-driven approach to estimate the take-over readiness of drivers in conditionally autonomous vehicles, based purely on the outputs of non-intrusive sensors observing the driver. Figure \ref{fig_pull} summarizes our approach. Our main contributions are as follows:

\begin{enumerate}
\item \textbf{Naturalistic dataset with autonomous driving:} We capture a 2 hour 10 min dataset of drivers behind the wheel of a commercially available conditionally autonomous vehicle.
This is captured using multiple high resolution cameras and depth sensors observing the driver. We use this data to train and evaluate our models. To the best of our knowledge, this is the first study evaluating take-over readiness of drivers using a naturalistic drive dataset from a conditionally autonomous vehicle.
\item \textbf{Human ratings for take-over readiness:} The goal of this work is to continuously estimate the take-over readiness of drivers using vision sensors. To test the feasibility of this approach, we collect ratings from multiple human raters viewing the sensor feed and analyze inter-rater agreement. Our experiments show a high consistency in the trend of the ratings across raters, rather than their absolute value. We normalize for rater bias using a percentile based approach. The mean value of the normalized ratings, averaged across raters, is then treated as the ground truth for our models. We term this the Observable Readiness Index (ORI).
\item \textbf{LSTM model for estimating take-over readiness:} We process the sensor streams frame by frame to analyze the drivers gaze \cite{vora2017generalizing}, pose \cite{cao2017realtime}, hand \cite{yuen2018looking,rangesh2018handynet} and foot activity, giving a holistic representation of the driver's state. We propose a Long Short Term Memory (LSTM) network to model the temporal dependency of the frame-wise representations. The LSTM continuously outputs the driver's ORI based on 2 seconds of past activity. Additionally, the model can recognize key-frames from the sensor streams that are most predictive of the driver's take-over readiness.
\end{enumerate}

\section{Related Research}

\subsection{Driver behavior analysis}
Driver behavior analysis based on vision sensors has been extensively addressed in prior research \cite{tawari2014robust,tawari2014driver,lee2011real,vasli2016driver,fridman2016driver,fridman2016owl,vora2017generalizing,vora2018driver,ohn2013vehicle,ohn2014beyond,ohn2014hand,borghi2018hands,rangesh2016hidden,deo2016vehicle,molchanov2015hand,yuen2018looking,yuen2016looking,rangesh2018handynet,tran2011pedal,tran2012modeling}. A large body of literature \cite{tawari2014robust,tawari2014driver,lee2011real,vasli2016driver,fridman2016driver,fridman2016owl,vora2017generalizing,vora2018driver} has focused on the driver's gaze estimation, being a useful cue for estimating the driver's attention. While early works \cite{lee2011real,tawari2014robust} have relied on head pose estimation for determining the driver's gaze, most recent approaches use a combination of head and eye cues \cite{tawari2014driver,vasli2016driver,fridman2016driver,fridman2016owl}. Recent work \cite{vora2018driver,vora2017generalizing} employing convolutional neural networks (CNNs) has allowed for generalizable estimation of driver gaze zones, across drivers and small variations in camera placement. Driver hand activity has also been the subject of a large number of research studies, being closely linked to the motor readiness of the driver. Common challenges involved in detecting the driver's hands such as fast varying illumination conditions, self occlusions, truncation have been outlined in \cite{das2015performance}. Many approaches have been proposed for detection, tracking and gesture analysis of the driver's hands while addressing some of these challenges \cite{ohn2013vehicle,ohn2014beyond,ohn2014hand,borghi2018hands,rangesh2016hidden,deo2016vehicle,molchanov2015hand}. Recent CNN models proposed by Yuen \textit{et al.} \cite{yuen2018looking} and Rangesh \textit{et al.}\cite{rangesh2018handynet} allow for accurate localization of the driver's hands in image co-ordinates and in 3-D respectively, and allow for further analysis such as hand activity classification and detection of objects held by the driver. A few works have also addressed driver foot activity analysis \cite{tran2011pedal,tran2012modeling}, being a complimentary cue to hand activity, for estimation of the driver's motor readiness. There has also been significant research that builds upon cues from driver gaze, hand and foot analysis for making higher level inferences such as driver activity recognition \cite{ohn2014head,braunagel2015driver,behera2018context}, driver intent prediction \cite{jain2015car,jain2016recurrent,martin2018dynamics,ohn2014predicting} and driver distraction detection \cite{liu2016driver,liang2007real,liang2014hybrid,bergasa2006real,li2015predicting,wollmer2011online}.

\subsection{Driver distraction estimation}

While very little literature exists on estimating the take-over readiness of drivers in autonomous vehicles, prior work \cite{liu2016driver,liang2007real,liang2014hybrid,bergasa2006real,li2015predicting,wollmer2011online} has addressed the closely related problem of driver distraction estimation in manually driven vehicles. Driver distraction estimation poses the same key challenges as take-over readiness estimation: defining a ground-truth metric for the quantity being estimated, and proposing models that map driver behavior to this metric. Here, we review the approaches used in prior work for addressing these challenges. Bergasa \textit{et al.} \cite{bergasa2006real} extract face and gaze features such as PERCLOS \cite{dinges1998perclos}, eye closure duration, blink frequency and face position and map them to the driver's vigilance level based on fuzzy rule expressions. Liang \textit{et al.} \cite{liang2007real,liang2014hybrid} and Liu \textit{et al.} \cite{liu2016driver} define the ground-truth for driver distraction as a binary variable, determined based on experimental conditions. The driver is considered distracted for trials involving the driver performing a secondary activity and not distracted for baseline trials, not involving secondary activities. Binary classifiers trained on features capturing the driver's head and eye activity and driving performance to detect driver distraction, with support vector machines (SVMs) used in \cite{liang2007real}, Bayesian networks used  in \cite{liang2014hybrid} and extreme learning machines (ELMs) used in \cite{liu2016driver}. Wollmer \textit{et al.} \cite{wollmer2011online} use subjective ratings provided by the drivers to define the ground truth distraction levels. They train an LSTM for classifying the ground truth distraction level using features based on driving performance and head pose of the driver. Li and Busso \cite{li2015predicting} use ratings provided by observers viewing the driver, rather than self evaluation by drivers. This allows for ratings that vary with time, and also allows for multiple raters to assign the rating. We use a similar approach for defining the ground-truth value for take-over readiness.

\subsection{Take-over time and quality studies}
A few recent studies \cite{gold2013take,braunagel2017ready,zeeb2015determines} have addressed the case of conditionally autonomous driving under simulator settings. Gold \textit{et al.} \cite{gold2013take} analyzed reaction times and take-over quality of drivers prompted with a take-over request 5 seconds and 7 seconds before a hazardous event, showing that drivers achieve faster reaction times for the 5 second interval, but poorer quality of take-overs. Zeeb \textit{et al.} \cite{zeeb2015determines} analyzed the relationship between driver gaze activity and take-over time and quality, showing that drivers preferring less frequent, longer glances at the secondary activity achieved slower take-overs, more prone to collisions, compared to drivers that preferred more frequent, but shorter glances. More recently, Braunagel \textit{et al.} \cite{braunagel2017ready} presented a model for classifying take-over readiness of drivers based on driver gaze and encodings of the driver's secondary activity and situation complexity. They defined the ground-truth for take-over readiness as a binary value, based on the quality of the performed take-overs. Our work differs from this approach on two counts; we evaluate our models using naturalistic drive data, and we train our models on more detailed descriptors of the driver's behavior.

\section{Experimental Setup}

\begin{figure}
\centering
\includegraphics[width=\columnwidth]{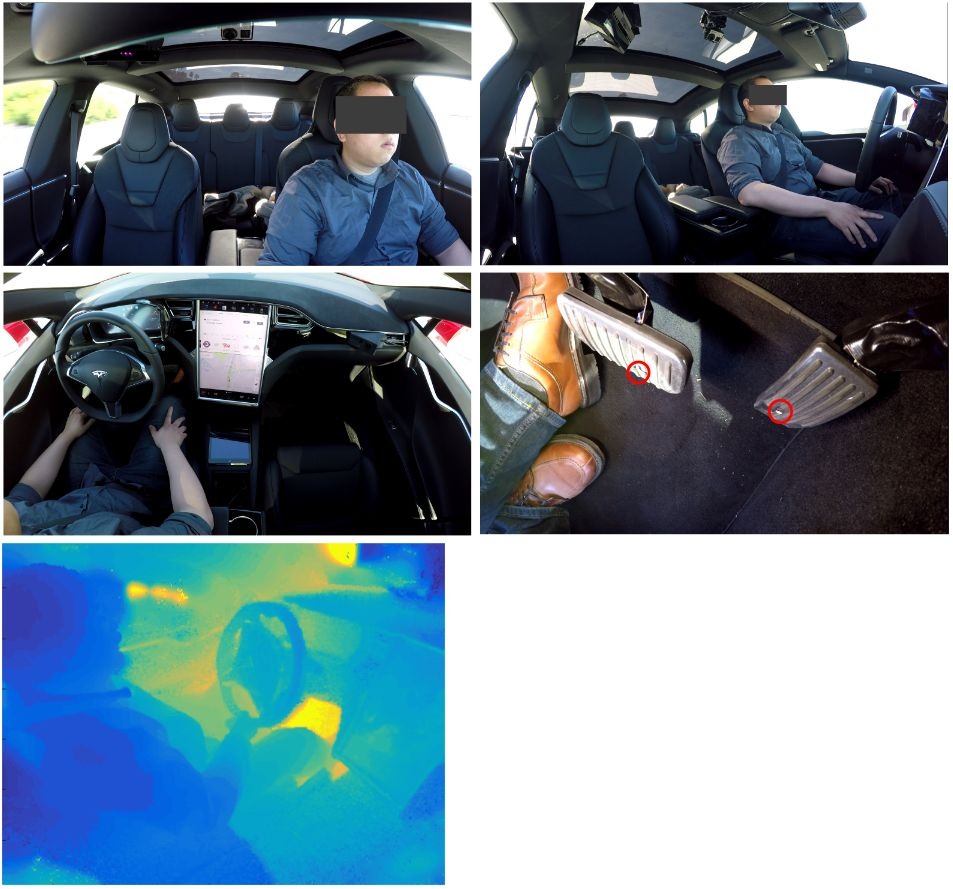}
\caption{\textbf{Experimental setup:} Our testbed is equipped with 4 high resolution cameras, a depth sensor and infrared sensors for foot pedals. This figure shows views used for driver face and gaze analysis \textbf{(top-left)}, hand activity analysis \textbf{(middle-left)}, pose analysis \textbf{(top-right)}, foot analysis \textbf{(middle-right)} with IR sensor locations, depth sensor output \textbf{(bottom)}}
\label{fig_testbed}
\end{figure}

We use our testbed \textbf{LISA-T} \cite{lisat}, built on top of a Tesla Model S, for our experiments. The vehicle has automatic cruise control and auto-steer capabilities for freeway driving. The testbed is equipped with a suite of non-intrusive sensors monitoring the driver's state. Four high resolution cameras observe the driver's face, hands, body pose and feet. These allow for computer vision algorithms to analyze the driver's gaze activity, hand activity, objects held and body posture. Additionally, a depth sensor observing the driver allows for analysis of the 3-D distances of the driver's hands from the vehicle controls. Figure \ref{fig_testbed} shows the camera views and depth sensor outputs. The testbed is also equipped with infrared sensors on its brake and gas pedals to measure the distance of the driver's foot from each pedal. All sensors are synchronized and record at a frame rate of 30 frames per second.

\section{Human ratings for observable driver readiness}

We collect subjective ratings provided by multiple human observers viewing video feed from our testbed, where the raters assign a value for the driver's readiness to take-over. The human ratings serve as a sanity check for our approach based on non-intrusive sensors. A high degree of agreement across raters would point to a measure for driver take-over readiness that could be estimated purely based on visual cues. Secondly, the ratings address the problem of ground-truth generation. The average rating provided by the raters can be used as the ground truth for training and evaluating a machine learning algorithm to estimate this measure of take-over readiness.

\begin{figure}
\centering
\includegraphics[width=\columnwidth]{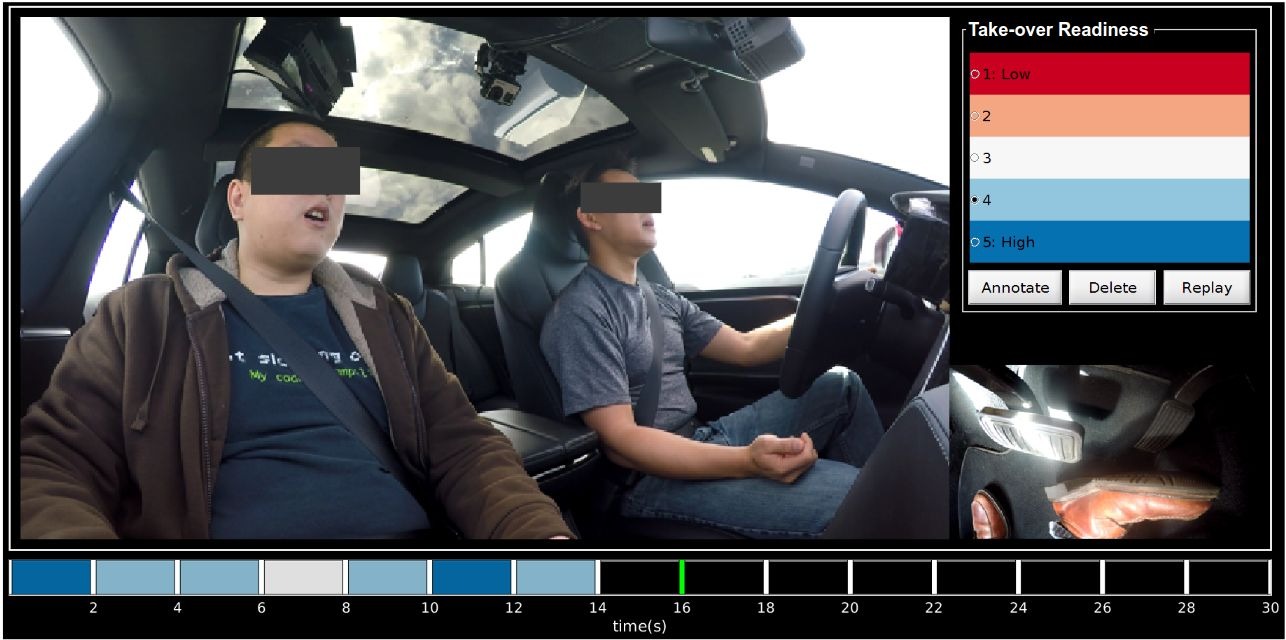}
\caption{\textbf{Interface for collecting ratings:} The raters observe video feed from the pose and foot cameras and assign a rating for each 2 second segment of video.}
\label{fig_tool}
\end{figure}

\subsection{Protocol for collecting ratings}

We chose a pool of 5 raters with driving experience and working knowledge of the Tesla autopilot system for assigning the subjective ratings. The raters were shown multiple 30 second video clips of drivers with the car driving in autonomous mode. Figure \ref{fig_tool} shows the interface used for collecting ratings. The raters were shown synchronized feed from the pose camera and foot camera. These two views were chosen in order to capture the complete state of the driver with the minimum number of views, so as to not overwhelm the raters.

The raters were given the following prompt:

\noindent\textit{"You will now be shown video clips of drivers behind the wheel of a vehicle operating in autonomous mode in freeway traffic. Rate on a scale of 1 to 5, with 1 being low and 5 being high, the readiness of the driver to take-over control from the vehicle for each 2 second video segment."}

\noindent We chose a discrete rating scale rather than a continuous scale to minimize rater confusion and expedite the rating process. We used discrete, fixed length time steps rather than flexible time boundaries for the same reason. The raters were allowed to replay the segments as well as update their previous ratings.

To prevent rater fatigue, in a single session, a rater rated a maximum of 25 clips, each 30 seconds long. The first two clips in every session were used to prime the rater for the task. The ratings for these two clips were discarded. The two priming clips were chosen such that one clip had primarily vigilant behavior, typically rated high by the raters, and one clip had the driver performing a distracting secondary activity, typically rated low.


\subsection{Dataset Description}
\begin{figure}
\centering
\includegraphics[width=\columnwidth]{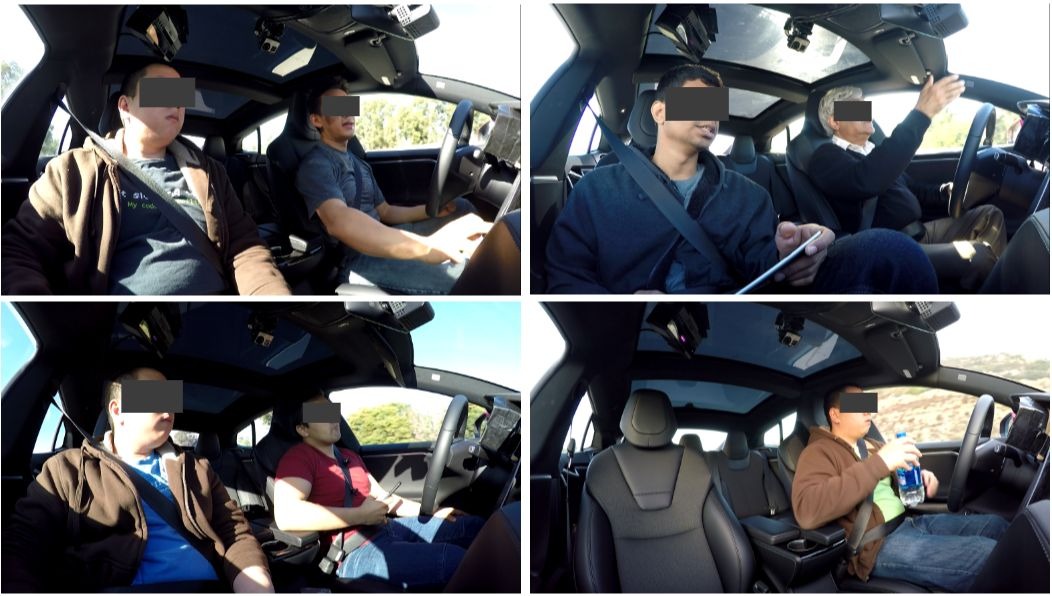}
\caption{\textbf{Examples frames from the dataset} (showing pose view): The dataset driver behaviors such as vigilant driving, talking to a co-passenger, gesturing, operating the infotainment unit, drinking a beverage and interacting with a cell-phone or tablet}
\label{fig_data}
\end{figure}

The complete dataset includes 260 clips, each 30 seconds long, amounting to 2 hours and 10 minutes of data. Among these clips, 20 were rated by the entire pool of 5 raters. We refer to this subset as the \textit{common set}. We use the common set to analyze the agreement across raters and to normalize for rater bias. The remaining 240 clips were rated by 2 different raters from the rater pool. We refer to this set as the \textit{expansion set}. We use both, the common set and the expansion set, for training and evaluating our models. The entire dataset involves naturalistic drives with 11 different drivers, with the car operating in autonomous mode on Californian multi-lane freeways. The data includes a wide range of driver behavior including vigilant driving, talking to a co-passenger, gesturing, operating the infotainment unit, drinking a beverage and interacting with a cell-phone or tablet. Figure \ref{fig_data} shows example frames from the collected dataset.

\subsection{Normalization of ratings}

One source of noise in the assigned ratings is rater bias. Raters can be strict or lax, and can use a varying range of values.  We normalize for rater bias using a percentile based approach. We use the common set for normalization of the ratings. We pool and sort ratings provided by each rater on the common set to obtain rater specific look-up tables. We then pool and sort ratings of all raters to obtain a combined look-up table. To normalize a specific rater's ratings, we find the percentile range of the assigned value in the rater's look-up table. We then replace it with the average of all values in that percentile range in the combined look-up table. This percentile based lookup can be applied to the entire dataset, including the expansion set.

\subsection{Observable Readiness Index}
We average the normalized ratings across raters to give a single rating for each 2 second interval in the clips. To obtain a continuously varying value representing the driver's take-over readiness, we interpolate these values over all frames using bi-cubic interpolation. We term this normalized and interpolated value the Observable Readiness Index (ORI), since it was assigned based purely on observable cues. We subsequently use the ORI as the ground truth for training and evaluating models for estimating the driver's take-over readiness.

\vspace{0.1in}

\subsection{Inter-rater agreement analysis}

We use intraclass correlation co-efficients (ICCs) as formulated by McGraw \textit{et. al.} \cite{mcgraw1996forming}, to evaluate inter-rater agreement. We model the human ratings as a two-way random-effect model without interaction, assuming $n$ observations and $k$ raters. Under this model, the rating $x_{ij}$ assigned by rater $j$ to clip $i$ can be expanded as,
\begin{equation}
x_{ij} = \mu + r_{i} + c_{j} + e_{ij},
\end{equation}
where, $\mu$ is the global average rating, $r_{i}$'s are the deviations based on the content of the rated clips, and $c_{j}$'s are the deviations due to rater bias. The $r_{i}$'s and $c_{j}$'s are independent, with mean 0 and variance $\sigma_{r}^{2}$ and $\sigma_{c}^{2}$ respectively.
And finally, $e_{ij}$ is the normally distributed measurement error with zero mean and variance $\sigma_{e}^2$

We report the following ICC values for the normalized and unnormalized ratings, as defined in \cite{mcgraw1996forming}:
\begin{itemize}

\item \textit{ICC(C,1):} This is given by the expression
\begin{equation}
ICC(C,1) = \frac{\sigma_r^2}{\sigma_r^2 + \sigma_e^2},
\end{equation}
and can be interpreted as the degree of consistency of the rating values. This is independent of the rater bias, and has a high value if the trend of ratings across raters is consistent.

\item \textit{ICC(A,1):} This is given by the expression
\begin{equation}
ICC(A,1) = \frac{\sigma_r^2}{\sigma_r^2 + \sigma_c^2 + \sigma_e^2}.
\end{equation}
This is the degree of absolute agreement of rater values, and has a high value only if the raters are in agreement in terms of the actual value of the ratings.

\item \textit{ICC(A,k):} This is given by the expression
\begin{equation}
ICC(A,k) = \frac{\sigma_r^2}{\sigma_r^2 + \frac{\sigma_c^2 + \sigma_e^2}{k}}.
\end{equation}
This can be interpreted as the reliability of the average rating provided by $k$ different raters. In our case, $k = 5$ for the common set, and $k = 2$ for the expansion set.
\end{itemize}

All ICC values are bounded between 0 and 1. The $\sigma$ values are estimated using two-way analysis of variances (ANOVA). Koo and Li \cite{koo2016guideline} prescribe that ICC values less than 0.5, between 0.5 and 0.75, between 0.75 and 0.9, and greater than 0.90 are indicative of poor, moderate, good, and excellent reliability, respectively.

\begin{table}
\caption{Rater agreement analysis based on intra-class correlation co-efficients (ICC)}
\label{tab_ira}
\begin{tabular*}{0.95\columnwidth}{@{}ccccc@{}}
\toprule
                                &                                 &                               &                               &                                        \\
\multirow{-2}{*}{Dataset}       & \multirow{-2}{*}{Normalization} & \multirow{-2}{*}{ICC(C,1)}    & \multirow{-2}{*}{ICC(A,1)}    & \multirow{-2}{*}{ICC(A,k)}             \\ \midrule
                                &                             & 0.584                         & 0.415                         & 0.780                                  \\
\multirow{-2}{*}{Common Set}    & \cellcolor[HTML]{EFEFEF} \ding{51}    & \cellcolor[HTML]{EFEFEF}0.580 & \cellcolor[HTML]{EFEFEF}0.582 & \cellcolor[HTML]{EFEFEF}\textbf{0.874} \\
                                &                             & 0.608                         & 0.517                         & 0.682                                  \\
\multirow{-2}{*}{Expansion Set} & \cellcolor[HTML]{EFEFEF} \ding{51}    & \cellcolor[HTML]{EFEFEF}0.637 & \cellcolor[HTML]{EFEFEF}0.627 & \cellcolor[HTML]{EFEFEF}\textbf{0.772} \\ \bottomrule
\end{tabular*}
\end{table}

Table \ref{tab_ira} shows the ICC values for the common and expansion sets, with and without normalization. As expected, the ICC(C,1) values are higher than the ICC(A,1) values due to the rater bias term $\sigma_{c}^2$ in the denominator for ICC(A,1). However, we note that normalization considerably improves the ICC(A,1) values for both the common and expansion sets, without affecting the ICC(C,1) values. This shows that the normalization maintains the trend ($\sigma_r^2$) of the ratings while reducing rater bias ($\sigma_c^2$). Finally, the last column shows the ICC(A,k) values, which represent the reliability of the average rating provided by all raters. As expected, this value is higher for the common set, rated by 5 raters as compared to the expansion set rated by 2 raters. However, both sets after normalization have an ICC(A,k) rating that falls within the range indicative of 'good reliability' of the ORI values as prescribed by Koo and Li \cite{koo2016guideline}.

\subsection{Qualitative analysis of ratings}

\begin{figure}
\centering
\includegraphics[width=\columnwidth]{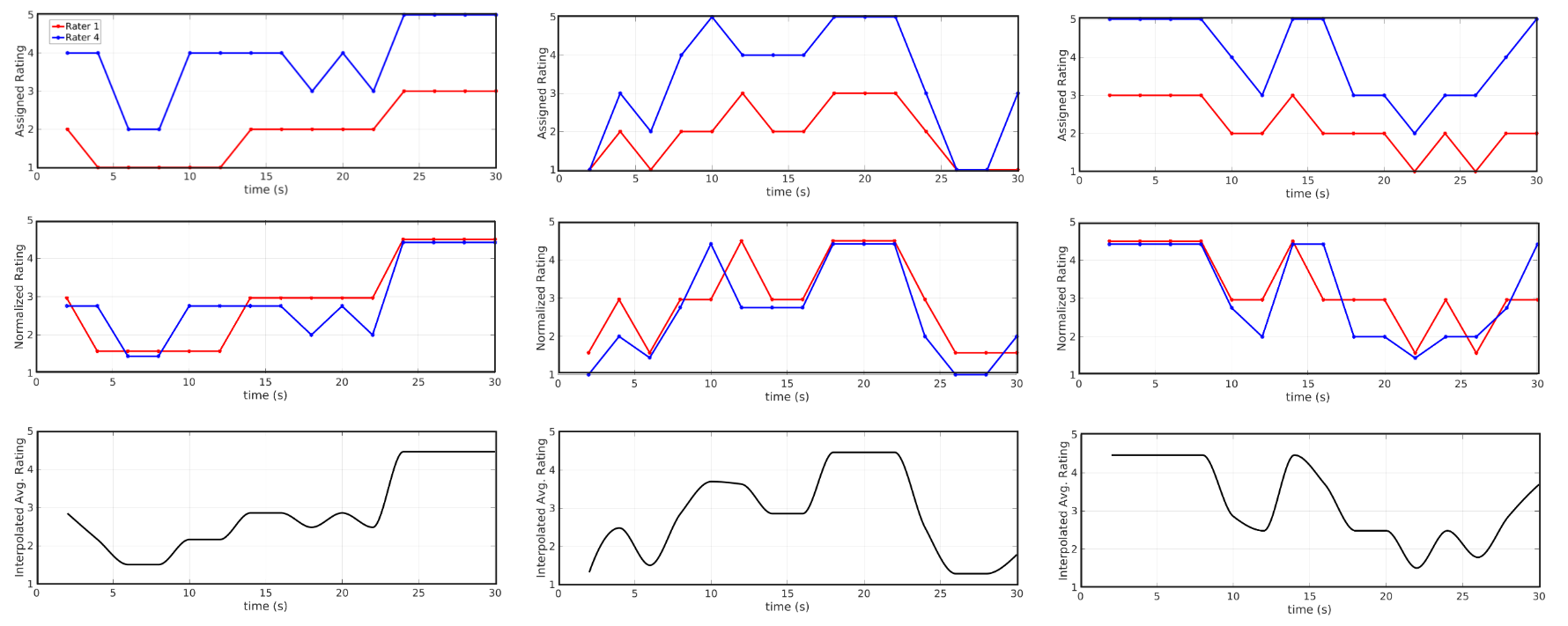}
\caption{\textbf{Example ratings:} Assigned (\textbf{top}), normalized, (\textbf{middle}) and averaged and interpolated (\textbf{bottom}) ratings provided by two raters for 3 sequences from the expansion set. The percentile based normalization scheme removes rater bias while retaining the trend of the ratings. Finally averaging and interpolating gives the continuously varying ORI for the sequences}
\label{fig_normalization}
\end{figure}
Figure \ref{fig_normalization} shows 3 examples of ratings from the expansion set. The top row shows the ratings assigned by raters 1 and 4. We observe that rater 1 is strict and rarely assigns a rating greater than 3. On the other hand, rater 4 is much more liberal, primarily assigning ratings from 3 to 5. However, we can observe the similarity in the trend of the ratings assigned by the two raters. The middle row shows the ratings after applying the percentile based normalization. We observe that normalization reassigns the ratings to a similar range of values while maintaining their trend, thereby removing rater bias.  Finally, the bottom row shows the continuously varying ORI, obtained after averaging and interpolation of the normalized ratings.

\section{Model for Estimating ORI}
\begin{figure*}[t]
\centering
\includegraphics[width=\textwidth]{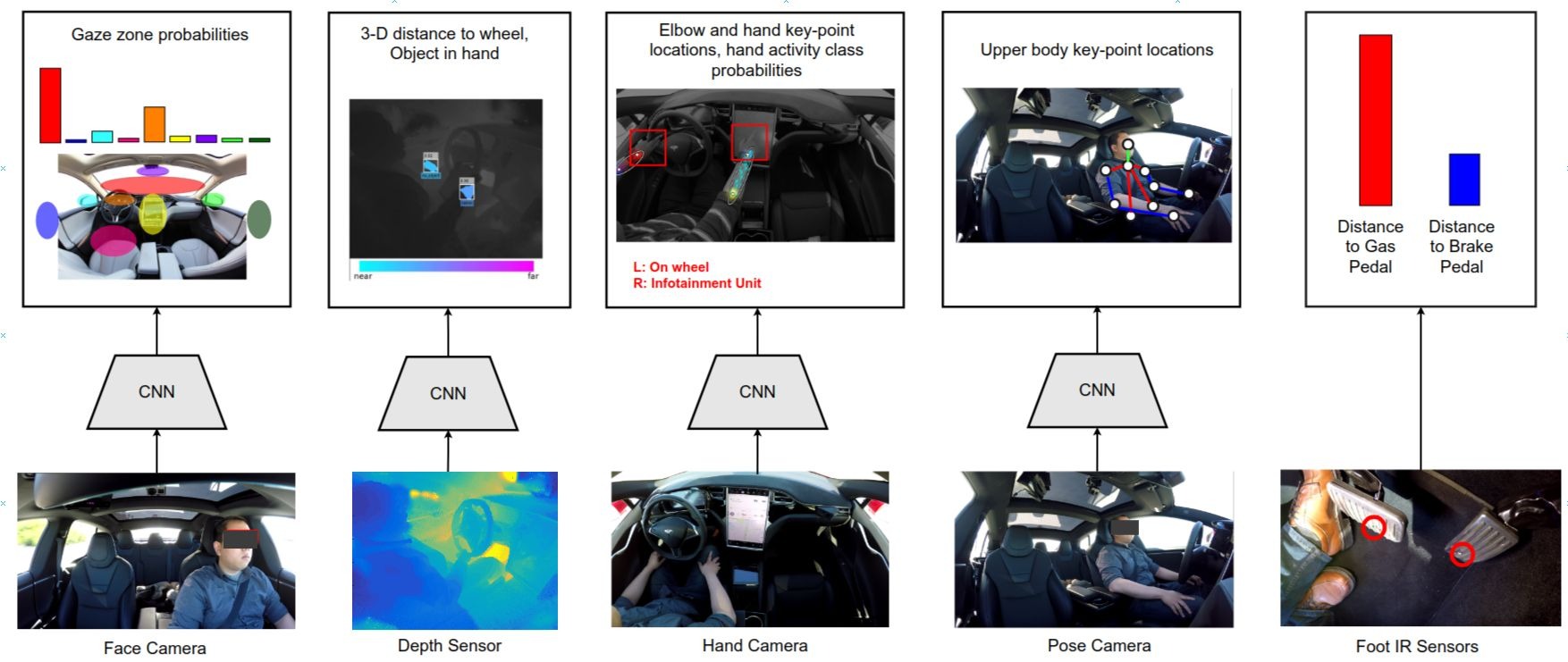}
\caption{\textbf{Frame-wise features capturing driver state:} We extract frame-wise features capturing driver's gaze, hand activity, pose and foot activity from the synchronized feed of our cameras and depth sensors. Existing convolutional neural network (CNN) based approaches \cite{vora2018driver,rangesh2018handynet,yuen2018looking,cao2017realtime} are used for extracting these frame-wise features. }
\label{fig_algo}
\end{figure*}
We wish to continuously estimate the driver's ORI based on recent feed from the sensors observing the driver. We pose this as a regression problem. At each instant, we treat the past two seconds of sensor feed as the input to our model and the ORI value as the output.

The raw sensor feed has high dimensionality due to the multiple high resolution cameras and depth sensors. To reduce the input dimensionality, we use existing models for driver behavior analysis \cite{vora2017generalizing,yuen2018looking,rangesh2018handynet,cao2017realtime} based on convolutional neural networks (CNNs). These models perform frame-wise analysis of the camera and depth sensor feed to extract features corresponding to driver gaze, hand activity, foot activity and pose. Figure \ref{fig_algo} summarizes the inputs and outputs of the frame-wise feature extraction models. Since all sensor streams are synchronized, the features can be concatenated to obtain a frame-by-frame representation of the driver's state.

While the CNNs extract a holistic frame-wise representation, driver activity spans across multiple frames. We thus need a model that can reason about the temporal dependencies of the features to estimate the ORI value. Due to their effectiveness at modeling long-term temporal dependencies in various sequence modeling tasks \cite{graves2013speech,graves2013generating,luong2014addressing,zaremba2014recurrent,donahue2015long}, we propose a model based on Long Short-Term Memory (LSTM) networks \cite{hochreiter1997long} for estimating ORI.

The following sections  describe the extracted frame-wise features and the proposed LSTM model.


\subsection{Frame-wise feature extraction}
\label{sec_feat}
\textbf{Gaze Analysis:}
We use the model proposed by Vora \textit{et al}. \cite{vora2017generalizing} for gaze analysis. The inputs to the model are frames from the face camera. The face detector described in \cite{yuen2016looking} is used for accurate localization of the driver's eye region. The cropped image of the driver's eye region is then passed through a CNN classifier, which outputs the driver's gaze zone. We consider 9 different gaze zones: \{\textit{Forward, left shoulder, left mirror, lap, speedometer, infotainment unit, rear-view mirror, right mirror, right shoulder}\} shown in Figure \ref{fig_algo}. The CNN uses a softmax output layer, giving frame-wise probabilities for each gaze zone. We use this 9 dimensional vector to represent driver gaze activity for each frame.

\textbf{Hand Analysis (Camera-based):} We use the model proposed by Yuen and Trivedi \cite{yuen2018looking}, for hand analysis based on feed from the hand camera. The model localizes the elbow and wrist joints of the driver using part affinity fields \cite{cao2017realtime}. The model is trained for robustness under rapidly changing lighting conditions and variability in hand posture during naturalistic drives. Additionally, the model
classifies hand activity for each hand using a CNN classifier. The CNN uses as input a cropped image near the localized wrist joints and outputs a hand-activity class. Four activity classes are considered for the left hand: \{\textit{On Lap, in air (gesturing), hovering wheel, on wheel}\}. Six classes are considered for the right hand: \{\textit{On Lap, in air (gesturing), hovering wheel, on wheel, interacting with infotainment unit, on cup-holder\}}. We obtain an 18 dimensional feature vector corresponding to the \textit{x} and \textit{y} co-ordinates of the 4 joints, the 4 dimensional output of the left hand activity classier and the 6 dimensional output of the right hand activity classifier for each frame of the hand camera.

\textbf{Hand Analysis (Depth-based):} We use HandyNet, proposed by Rangesh and Trivedi \cite{rangesh2018handynet}, for hand analysis using the depth sensor. The model performs instance segmentation on each frame of the depth sensor to output the locations of the driver's hands. This allows for computation of the distance of the driver's hands from the vehicle controls in 3-D. Additionally, the model also classifies the object held by the driver, if any. We consider 4 different object classes: \{\textit{No object, cell-phone/tablet, beverage/food, other item}\}. Thus, for each frame of the depth sensor, we obtain a 5 dimensional feature vector, corresponding to the distance to the wheel, and the 4 dimensional output of the hand object classifier.

\textbf{Pose Analysis:} The driver's body posture can be a useful additional cue for estimating their readiness to take over. We capture the driver's pose for each frame by using OpenPose \cite{cao2017realtime} proposed by Cao \textit{et al}. The model is capable of localizing 18 body key-point locations. For our task, we only consider the the 10 upper body key-points visible in the pose camera view, as shown in Figure \ref{fig_algo}. The \textit{x} and \textit{y} co-ordinates of the 10 body key-points give a 20 dimensional feature vector for each frame of the pose camera.

 \textbf{Foot Analysis:} To obtain a representation of the driver's foot activity, we use the outputs of the IR pedal sensors. These give the distance of the driver's foot to the gas and brake pedal for each frame.

\begin{figure}
\centering
\includegraphics[width=\columnwidth]{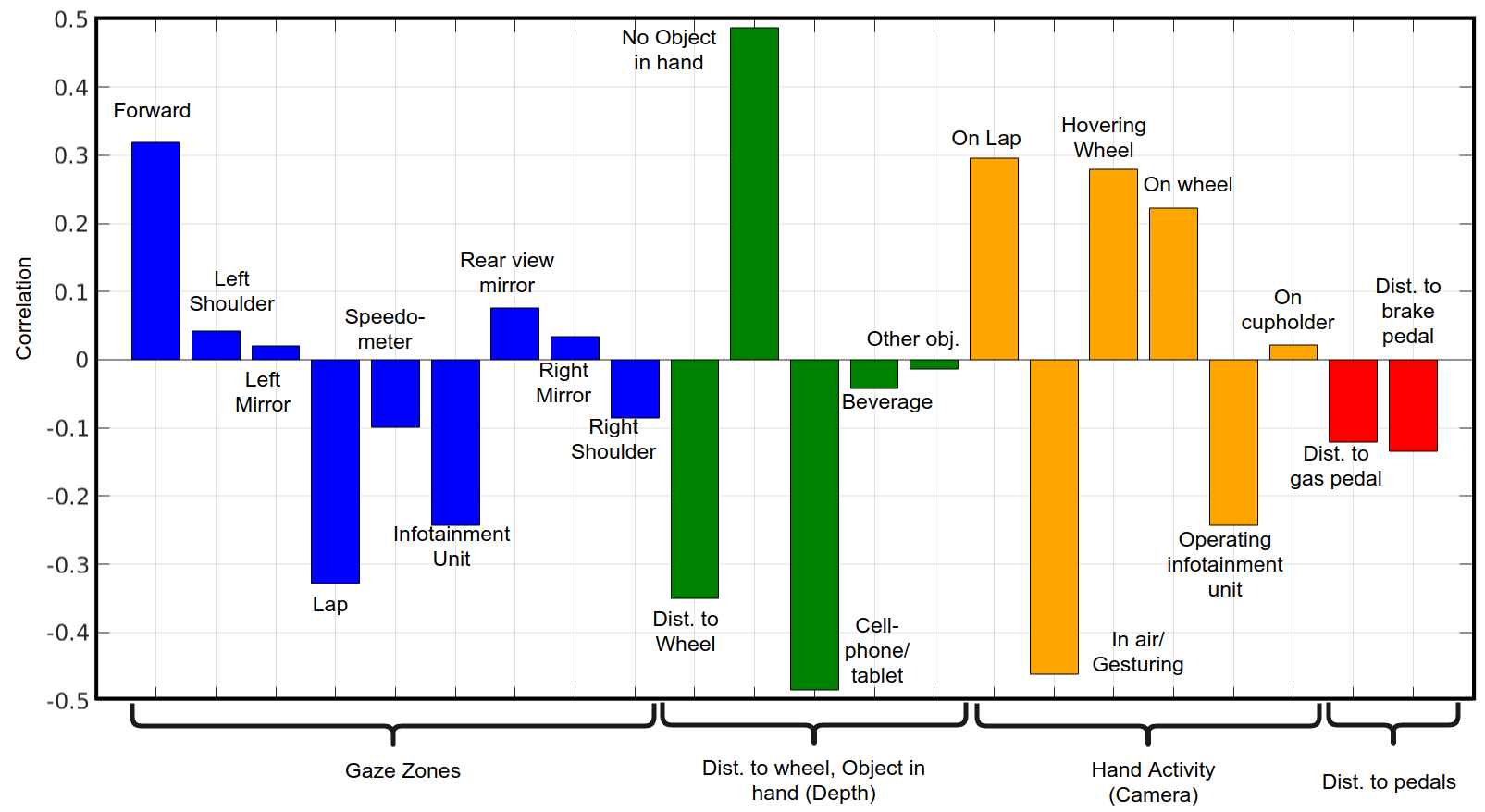}
\caption{Frame-wise correlation of gaze, hand and foot features with ORI ratings}
\label{fig_corr}
\end{figure}



\subsection{Correlation of extracted features with ORI}
Figure \ref{fig_corr} shows the frame-wise correlation of features with the ORI values. We note that in general, hand features seem to have higher correlation with ORI compared to gaze or foot features. In particular, the presence of a hand-held phone or tablet has a strong negative correlation with the rated ORI values. On the other hand, the driver's hands being free ( without holding any object) has a high positive correlation with ORI. In terms of activity classes, gesturing and interacting with the infotainment unit are negatively correlated with ORI, whereas the hands hovering or being on the wheel are positively correlated. Although the gaze features have a lower correlation with ORI as compared to hand features, we note that the gaze zones corresponding to sensing the vehicle's surroundings such as looking forward or viewing the mirrors are correlated positively with ORI. On the other hand, gaze zones corresponding to driver distraction such as looking at the lap (or hand-held device), infotainment unit or speedometer have negative correlation. The only exception to this seems to be the right shoulder gaze zone, which is negatively correlated with ORI. This could be because the driver may look to the right to interact with the co-passenger in the front seat, which can be seen as a distracting activity. Finally, we note that the distances of the hands and feet from the vehicle controls are negatively correlated with the rated ORI, which is reasonable since these affect the motor-readiness of the driver.


\subsection{Proposed LSTM model}
In its simplest form, we can define an LSTM model for processing the sequence of frame-wise features as shown in Figure \ref{fig_a}. An LSTM is a recurrent neural network. It updates its state at each time-step based on its state at the previous time-step and the frame-wise features at the current time-step. The final state of the LSTM can be expected to encode the driver's activity spanning over 2 seconds of context by processing the temporal dependencies of the frame-wise features. The final state of the LSTM is then processed by an output layer to give the scalar valued ORI for the 2 second snippet of driver activity.

\begin{figure}
    \begin{subfigure}{0.29\columnwidth}
    	\label{fig_a}
        \centering
        \includegraphics[height=1.2in]{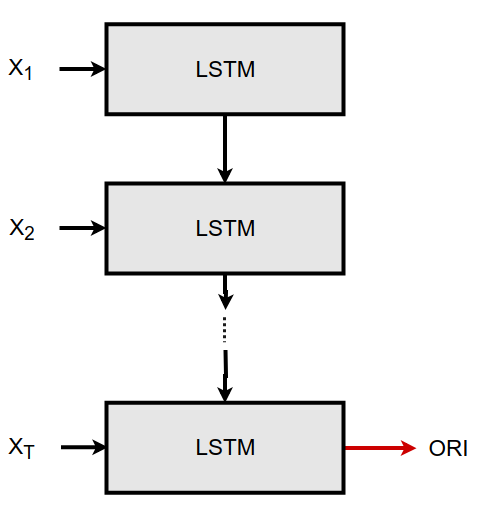}
        \caption{Simple LSTM}
    \label{fig_a}
    \end{subfigure}%
    ~
    \begin{subfigure}{0.7\columnwidth}
        \centering
        \includegraphics[height=1.2in]{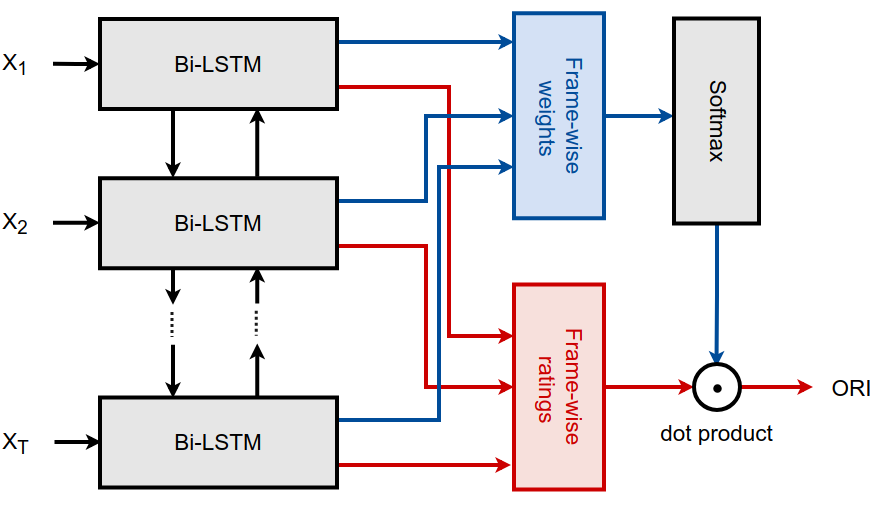}
        \caption{LSTM with key-frame weighting}
    \label{fig_b}
    \end{subfigure}
    \caption{LSTM models used for estimating ORI}
\end{figure}

\subsubsection*{Key-frame weighting} While the LSTM reasons over the complete 2 second sequence of frame-wise features, driver activity in certain parts of the sequence can be more strongly predictive of their readiness to take over. For example, a driver may for a split second, interact with their phone or the infotainment unit, while otherwise appearing vigilant in terms of hand, eye and foot activity. While this activity may not span 2 seconds, it is indicative of driver distraction and low readiness to take over. Human raters viewing the driver would take this into consideration and weight this part of the video clip more strongly for assigning their ratings. We thus need a model capable of assigning greater weight to key-frames in the sequence of features, to determine the final rating.

We propose the model shown in Figure \ref{fig_b} for key-frame weighting. The LSTM outputs two values at each time step, the rating for that frame and the weight for that frame. The frame-wise weights are normalized using a softmax layer. A dot product of the normalized weights and the frame-wise ratings gives the ORI value for the sequence. This allows the model to assign a greater weight to key-frames in the sequence of features. We use a bidirectional LSTM \cite{graves2005framewise} instead of a unidirectional LSTM. This allows for information to flow from the entire 2 second sequence to LSTM states at each time step. This allows the model to relatively weigh the current frame with respect to the other frames in the sequence in order to determine the frame-wise weight. Our approach can be considered similar to the attention mechanism \cite{bahdanau2014neural} in LSTM encoder-decoder models, with an attention based decoder applied for a single step.

\subsubsection*{Implementation details}
We use bi-directional LSTMs with a 32 dimensional state vector for both the forward and backward LSTMs. We use a sigmoid activation for the frame-wise ratings, giving a predicted ORI value between 0 and 1. The ground truth ORI ratings ranging from 1 to 5 are shifted and scaled to map them to the 0 to 1 range. We train the model by minimizing mean absolute error on the train set. We use Adam \cite{adam} with learning rate 0.001 to train the model. The models are trained using Pytorch \cite{pytorch}.

\section{Experimental Evaluation}

For our experiments, we split our dataset of 260 video clips, into a train set, validation set and a test set. The train set consists of 172 video clips from 22 different drives, the validation set consists of 32 video clips from 5 drives and the test set consists of 56 clips from 6 drives. The train, validation and test sets were selected from different drives. The test set includes data from 6 different drivers, 3 of whom were not present in the train or validation sets. Since each clip is 30 seconds long, with the sensors recording at 30 fps, this gives us about 154,000 training samples, 28,000 validation samples and 50,000 test samples. We train our models to minimize the mean absolute error between the predicted and assigned ORI values on the train set. We select the model with the minimum mean absolute error on the validation set for reporting results on the test set.

We report the mean absolute error (MAE) between the predicted and assigned ORI values for the test set. We compare the following models:
\begin{itemize}
\item \textit{Frame-wise SVM baseline:} We use a linear SVM trained purely using the frame-wise features as the simplest baseline. This allows us to evaluate the effect of using temporal context for estimating ORI.
\item \textit{Simple LSTM:} We also report results using a simple LSTM as shown in Figure \ref{fig_a}. For a fair comparison with our proposed model using bidirectional LSTMs, we use LSTMs with 64 dimensional states for the simple LSTM, while using LSTMs with 32 dimensional states for the forward and backward components of the bidirectional LSTM.
\item \textit{LSTM with key-frame weighting:} Finally, we report results using our proposed model with key-frame weighting shown in Figure \ref{fig_b}.
\end{itemize}
Additionally we perform ablations with the feature streams described in Section \ref{sec_feat} to analyze the importance of gaze, hand, pose and foot features for estimating the driver's ORI.

Table \ref{tab_results} shows the MAE values for the three models compared, trained using one or more feature streams. We note that the LSTM models achieve a significantly lower MAE than the frame-wise baseline for most combinations of feature streams, showing that modeling temporal context is useful for estimating the ORI. We also note that key-frame weighting consistently leads to lower errors for different features, compared to the simple LSTM. The top four rows of Table \ref{tab_results} shows the MAE values for models trained on a single feature stream of gaze, hand, pose or foot features. All models achieve an MAE of within 1 point of the average rating assigned by the human raters on the 5 point scale. This shows that each feature stream has useful cues for estimating ORI. Comparing the features, we see that the hand features seem to have the most useful cues for estimating ORI, closely followed by the gaze features. The last three rows of the table show the MAE values achieved by combining the feature streams. We note that combining the hand and gaze features significantly lowers the MAE compared to single feature models. The pose and foot features lead to further improvements, giving the lowest MAE value of 0.449.

\begin{table}[]
\caption{Mean absolute error (MAE) of predicted ORI values with respect to assigned values}
\label{tab_results}
\begin{tabular*}{\columnwidth}{@{}ccccccc@{}}
\toprule
\multicolumn{4}{c}{Features used}                 & \multirow{2}{*}{\begin{tabular}[c]{@{}c@{}}Frame-wise\\ SVM\end{tabular}} & \multirow{2}{*}{\begin{tabular}[c]{@{}c@{}}Simple\\  LSTM\end{tabular}} & \multirow{2}{*}{\begin{tabular}[c]{@{}c@{}}LSTM \slash w key-\\ frame wts\end{tabular}} \\ \cmidrule(r){1-4}
Gaze       & Hand       & Pose       & Foot       &                                                                           &                                                                         &                                                                                             \\ \midrule
\ding{51} &            &            &            & 0.779                                                                         & 0.621                                                                   &\textbf{ 0.581}                                                                                       \\
\cellcolor[HTML]{EFEFEF}           & \cellcolor[HTML]{EFEFEF}\ding{51} &\cellcolor[HTML]{EFEFEF}            &\cellcolor[HTML]{EFEFEF}            &\cellcolor[HTML]{EFEFEF} 0.639                                                                         &\cellcolor[HTML]{EFEFEF} \textbf{0.571}                                                                   & \cellcolor[HTML]{EFEFEF}0.572                                                                                       \\
           &            & \ding{51} &            & 0.836                                                                          & 0.855                                                                   & \textbf{0.823}                                                                                       \\
\cellcolor[HTML]{EFEFEF}           & \cellcolor[HTML]{EFEFEF}           & \cellcolor[HTML]{EFEFEF}           & \cellcolor[HTML]{EFEFEF}\ding{51} & \cellcolor[HTML]{EFEFEF}\textbf{0.986}                                                                         & \cellcolor[HTML]{EFEFEF}1.018                                                                   & \cellcolor[HTML]{EFEFEF}1.001                                                                                       \\
\ding{51} & \ding{51} &            &            & 0.611                                                                         & 0.470                                                                   & \textbf{0.457}                                                                                       \\
\cellcolor[HTML]{EFEFEF}\ding{51} & \cellcolor[HTML]{EFEFEF}\ding{51} & \cellcolor[HTML]{EFEFEF}\ding{51} &   \cellcolor[HTML]{EFEFEF}         & \cellcolor[HTML]{EFEFEF}0.602                                                                         & \cellcolor[HTML]{EFEFEF}0.468                                                                   & \cellcolor[HTML]{EFEFEF}\textbf{0.456}                                                                                       \\
\ding{51} & \ding{51} & \ding{51} & \ding{51} & 0.599                                                                         & 0.467                                                                   & \textbf{0.449}                                                                              \\ \bottomrule
\end{tabular*}
\end{table}

\begin{figure*}
\centering
\includegraphics[width=0.9\textwidth]{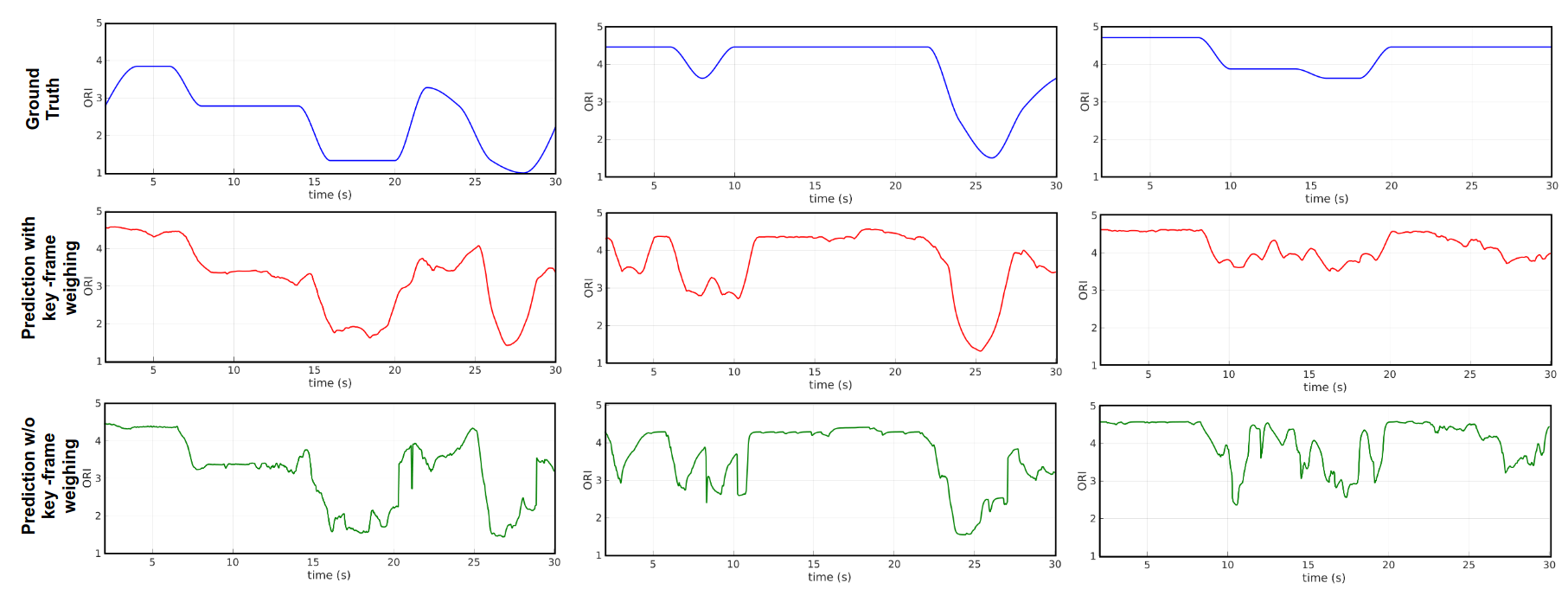}
\caption{\textbf{Effect of key-frame weighting model:} Three example clips with ground truth ORI (top), ORI predicted by simple LSTM (bottom), ORI predicted with key-frame weighting (middle). Key-frame weighting allows the model to focus on the most relevant frames in the sequence and generate a smoother, more reliable rating, compared to the noisier, more reactive simple LSTM.}
\label{fig_ex1}
\end{figure*}

\begin{figure*}
\centering
\includegraphics[width=0.9\textwidth]{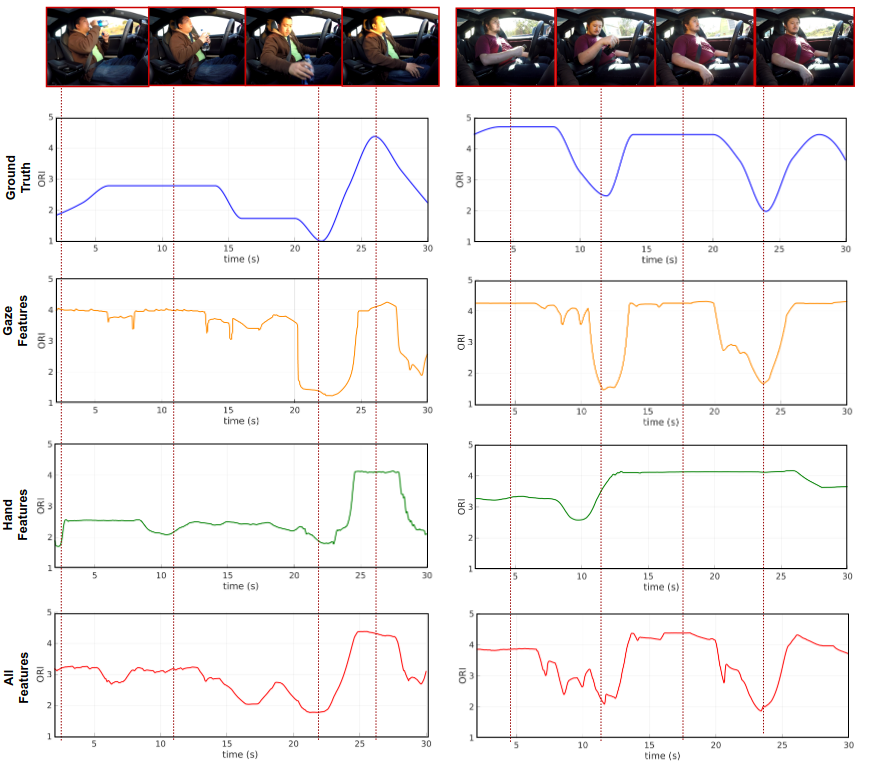}
\caption{\textbf{Importance of gaze and hand cues:} This figure shows two example clips with ratings predicted based purely on gaze cues, hand cues and all features combined. The first example shows a case where the gaze features fail to correctly predict the ORI, where the hand features can be used to correct the error. The second example shows a failure case of hand features, that could be corrected based on the gaze features. The model trained on the combined feature streams correctly predicts the ground truth rating for both cases.}
\label{fig_ex2}
\end{figure*}

\subsection{Effect of key-frame weighting model}
We qualitatively analyze the ratings produced by a simple LSTM and our proposed model.
Figure \ref{fig_ex1} shows ratings for three example clips. The top row (blue) shows the ground truth ratings, provided by the raters. The middle row (red) shows ratings predicted by the model with key-frame weighting. The bottom row (green) shows ratings predicted by a simple LSTM model. We observe that while both models manage to predict the trend of the ground truth ratings, the simple LSTM produces a noisy estimate, rapidly varying in short intervals. On the other hand the model with key-frame weighting produces a much smoother rating. This can be explained by the sparsity of key-frames in the sequence. Since the model primarily relies on a subset of the most important sections of the sequence to assign an ORI rating, it leads to a smoother, more reliable output throughout the sequence compared to the simple LSTM, which is more reactive.

\subsection{Ablation of feature streams}

We also compare the ratings generated by models trained purely using the gaze or hand features, with those generated by the combined feature stream model to analyze the importance of each cue for estimating the ORI. Figure \ref{fig_ex2} shows the ratings for two example clips.  The top row (blue) shows the ground truth ratings, provided by the raters. The second (yellow) and third (green) rows show ratings predicted based purely on gaze and hand features respectively. Finally, the bottom row (red) shows the ratings predicted by the model trained using all feature streams. Additionally, we also show snapshots from the pose camera at different instants to provide context.

The first example shows the driver drinking water from a bottle. For the interval from 0-20 seconds in the clip, the driver has the bottle in their hand. However, they have their eyes on the road. We note that the gaze model overestimates the ORI value for this interval, while the hand model underestimate the ORI. The rating from the combined model closely matches the ground truth value. We note that all models correctly estimate the ORI for the last two instants shown, at 22 seconds and 26 seconds. At 22 seconds, the driver has a bottle in their hand while looking away from the road to put it away. All models assign a low value for this instant. At 26 seconds, the driver is attentive, with their hands close to the wheel. All models correctly assign a high rating at this instant.

The second example shows the driver viewing the screen of a hand held device, and later viewing the screen of the infotainment unit. We see that both events correspond to dips in the ORI ratings assigned by the raters. We observe that the hand model correctly assigns a low rating for the first case, due to the presence of a hand held device in the driver's hands. However, it fails to capture the dip in ORI due to the driver glancing at the infotainment unit. The gaze and combined models, on the other hand correctly predict both dips in the ORI value.

\section{Conclusions}

In this study, we proposed an approach for estimating the take-over readiness of 'drivers' in autonomous vehicles based on non-intrusive vision sensors observing them. Our approach was evaluated using an extensive naturalistic drive dataset captured in a commercially available conditionally autonomous vehicle.

We collected subjective ratings for driver take-over readiness from observers viewing the sensor feed. Analysis of the assigned ratings in terms of intra-class correlation coefficients showed high consistency in the ratings assigned by the raters, suggesting that human raters agree upon an observable measure for the driver's take-over readiness. We normalized the ratings for rater bias and averaged across raters to extract this measure, which we termed the observable readiness index (ORI).

We presented a model for estimating the ORI, based on CNNs for extracting frame-wise representations of the driver's gaze, hand , pose and foot activity and an LSTM for learning their temporal dependencies. Our best model achieves an MAE of 0.449 on the five point rating scale, showing that the ORI can be reliably estimated. Ablation of feature streams shows that each feature stream is useful for estimation of th ORI, with the order of importance being hand features, gaze features, pose features and finally foot features. Finally, we proposed a modification to simple LSTMs, to allow for detecting key-frames in the input sequence most predictive of the driver's take-over readiness. Our experiments show that this leads to lower MAE of the predicted ORI, and leads to a smoother, more reliable prediction compared to the noisier, more reactive prediction from a simple LSTM.

\color{black}



\section*{Acknowledgments}
We would like to thank the raters who participated in this study for their
valuable time, and our colleagues at the Laboratory for
Intelligent and Safe Automobiles for their help with setting up the test bed and for data collection. We thank our industry sponsors at the Toyota Collaborative Safety Research Center (CSRC) for their continued support.

\ifCLASSOPTIONcaptionsoff
  \newpage
\fi

{\small
\bibliographystyle{IEEEtran}
\bibliography{egbib}
}

\newpage
\begin{IEEEbiography}[{\includegraphics[width=1in,height=1in,clip,keepaspectratio]{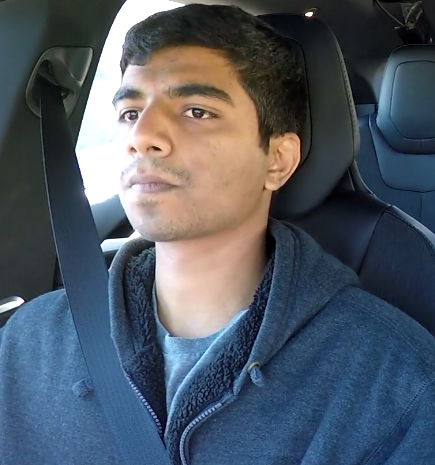}}]{Nachiket Deo}
is currently working towards his PhD in electrical engineering from the University of California at San Diego (UCSD), with a focus on intelligent systems, robotics, and control. His research interests span computer vision and machine learning, with a focus on motion prediction for vehicles and pedestrians
\end{IEEEbiography}
 \vspace{-120 mm}
\begin{IEEEbiography}[{\includegraphics[width=1in,height=1.25in,clip,keepaspectratio]{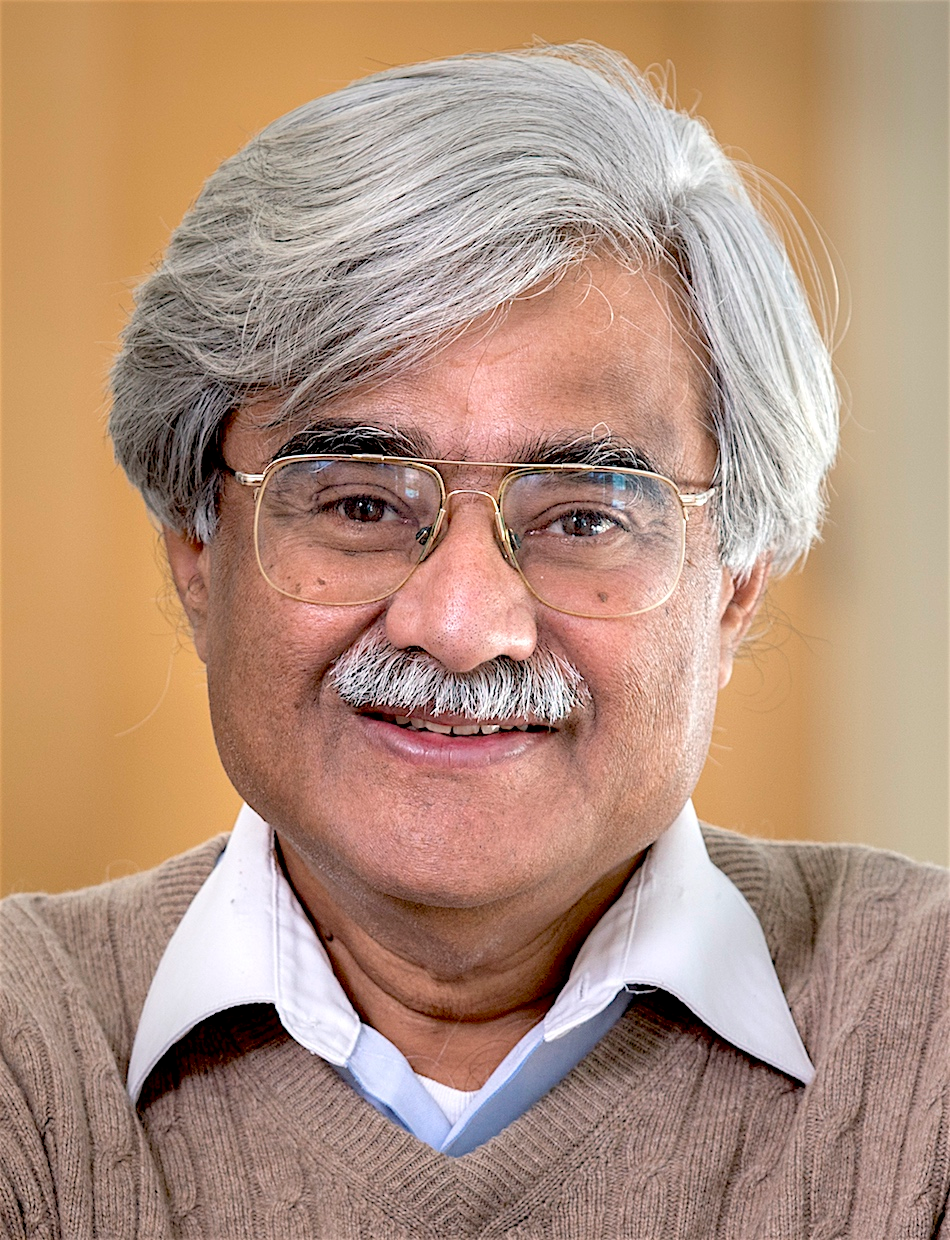}}]{Mohan Manubhai Trivedi}
is a Distinguished Professor at University of California, San Diego (UCSD) and the founding director of the UCSD LISA: Laboratory for Intelligent and Safe Automobiles,
winner of the IEEE ITSS Lead Institution Award (2015). Currently, Trivedi and his team
are pursuing research in intelligent vehicles, autonomous driving, machine perception, machine learning, human-robot interactivity, driver assistance. Three of his students have received "best dissertation" recognitions and over twenty best papers/finalist recognitions. Trivedi is a Fellow of IEEE, ICPR and SPIE. He received the IEEE ITS Society's highest accolade "Outstanding Research Award" in 2013. Trivedi serves frequently as a consultant to industry and government agencies in the USA and abroad.
\end{IEEEbiography}

\end{document}